\documentclass[lettersize,journal]{IEEEtran}
\usepackage{amsmath,amsfonts}
\usepackage{algorithmic}
\usepackage{algorithm}
\usepackage{array}
\usepackage{enumitem}
\usepackage[caption=false,font=normalsize,labelfont=sf,textfont=sf]{subfig}
\usepackage{textcomp}

\usepackage{stfloats}
\usepackage{url}
\usepackage{verbatim}
\usepackage{graphicx}
\usepackage{cite}
\usepackage{booktabs}
\usepackage{multirow}
\usepackage{amsmath}
\usepackage{colortbl}
\usepackage{xcolor, soul}
\usepackage{hyperref}
\usepackage{subcaption}
\usepackage{cleveref}

\usepackage{array}
\usepackage{tikz}
\definecolor{level1}{RGB}{0,0,0}

\tikzset{
    arrow1/.style={
        thick,
        color=level1,
        shorten >=1pt,
        shorten <=1pt
    }
}

\newcommand{\conversation}[1]{%
    \begin{tabular}[t]{@{}l@{}}
        #1
    \end{tabular}%
}

\newcommand{\name}{\textsc{CrowdShield}}
\newcommand{\dataset}{\textsc{MisT}}

\begin{document}

\title{Crowd Intelligence for Early Misinformation Prediction on Social Media}

\author{Megha Sundriyal$^1$, Harshit Choudhary$^1$, Tanmoy Chakraborty$^2$, Md Shad Akhtar$^1$
\\
\textit{$^1$Indraprastha Institute of Information Technology Delhi, India, $^2$Indian Institute of Technology Delhi, India}}

\markboth{Journal of \LaTeX\ Class Files,~Vol.~14, No.~8, August~2021}%
{Shell \MakeLowercase{\textit{et al.}}: A Sample Article Using IEEEtran.cls for IEEE Journals}

\maketitle

\begin{abstract}
Misinformation spreads rapidly on social media, causing serious damage by influencing public opinion, promoting dangerous behavior, or eroding trust in reliable sources. It spreads too fast for traditional fact-checking, stressing the need for predictive methods. We introduce \name, a crowd intelligence-based method for early misinformation prediction. We hypothesize that the crowd's reactions to misinformation reveal its accuracy. Furthermore, we hinge upon exaggerated assertions/claims and replies with particular positions/stances on the source post within a conversation thread. We employ Q-learning to capture the two dimensions -- stances and claims. We utilize deep Q-learning due to its proficiency in navigating complex decision spaces and effectively learning network properties. Additionally, we use a transformer-based encoder to develop a comprehensive understanding of both content and context. This multifaceted approach helps ensure the model pays attention to user interaction and stays anchored in the communication's content. We propose \dataset, a manually annotated misinformation detection Twitter corpus comprising nearly $200$ conversation threads with more than $14K$ replies. In experiments, \name\ outperformed ten baseline systems, achieving an improvement of $\sim 4\%$ macro-F1 score. We conduct an ablation study and error analysis to validate our proposed model's performance. The source code and dataset are available at \url{https://github.com/LCS2-IIITD/CrowdShield.git}

\end{abstract}


\section{Introduction}
\IEEEPARstart{T}{HE} proliferation of misinformation over social media platforms has become a critical challenge. 
The shear amount of misinformation across platforms significantly impacts public opinion and can have tangible real-world consequences. 
 
Notable events such as the 45th Presidential US Elections, the COVID-19 pandemic, and the Ukraine-Russia war highlight the adverse impact of online misinformation on society \cite{caceres2022impact, nisbet2021presumed, chuai2024topic}. Manual fact-checkers struggle to identify misinformation due to the enormous volume of content generated and shared on social media daily. As a result, researchers have begun investigating automated methods addressing this problem \cite{sun2023inconsistent, liao2021integrated, schlicht2024automatic, gupta2021lesa}. 

\textbf{\textit{Early prediction.}} Moreover, the rapid spread of misinformation on social media poses another formidable challenge in containing the misinformation early before it reaches a wide audience. 
By identifying misinformation before it gains substantial traction, we can mitigate its negative effects and uphold the integrity of online discourse. The concept of early detection has been well-studied for online information disorders such as hate speech \cite{lin2021early}, rumors \cite{zhou2019early}, fake news \cite{liu2018early}, etc. 
At the same time, developing robust methods to detect misinformation early is also essential to protect the accuracy of information and promote meaningful discussions. 
Recently, there have been major advancements in the early detection of misinformation \cite{liu2018early, chen2018call}. These efforts aim to strengthen digital platforms against the harmful effects of misinformation and ensure that accurate and reliable information prevails online. In this work, we aim to devise an approach for early misinformation detection in social media utilizing \textit{crowd intelligence}.

\textbf{\textit{Limitations of existing studies.}} Conventional approaches to discern misinformation typically rely on linguistic patterns or external knowledge sources to ascertain whether the information is true. Content-based detection hinges heavily on linguistic cues, emotions, or sentiments \cite{castillo2011information, qazvinian2011rumor}. However, these methods come with certain shortcomings. Firstly, the messages on social media sites like Twitter and Facebook are short and informal. As a result, linguistic features extracted from them tend to be insufficient for deep learning algorithms. Moreover, they often use excessive informal text and slang, making it challenging to follow proper grammar and syntax. 
Another line of research concentrates on evidence-based misinformation identification \cite{yang2024towards, hangloo2023evidence, gad2019exfakt}. These systems are generally reliable and depend on established evidence to validate the posts. However, these approaches can be time-consuming and unable to keep up with the swift propagation of misinformation. At the same time, they often lack sufficient evidence to verify newly posted information. Recent studies have sought innovative methods to detect misinformation that combine temporal features extracted from user responses and propagation networks \cite{fang2023unsupervised, wu2024graph, wu2023adversarial}. We also hypothesize that the \textit{`crowd intelligence'} through these user responses hold a wealth of direct or indirect cues, which can be leveraged to assess the credibility of the source posts effectively.

\begin{figure*}[!th]
\centering
\subfloat[\footnotesize{Non-misinformation}]{\includegraphics[width=0.45\textwidth]{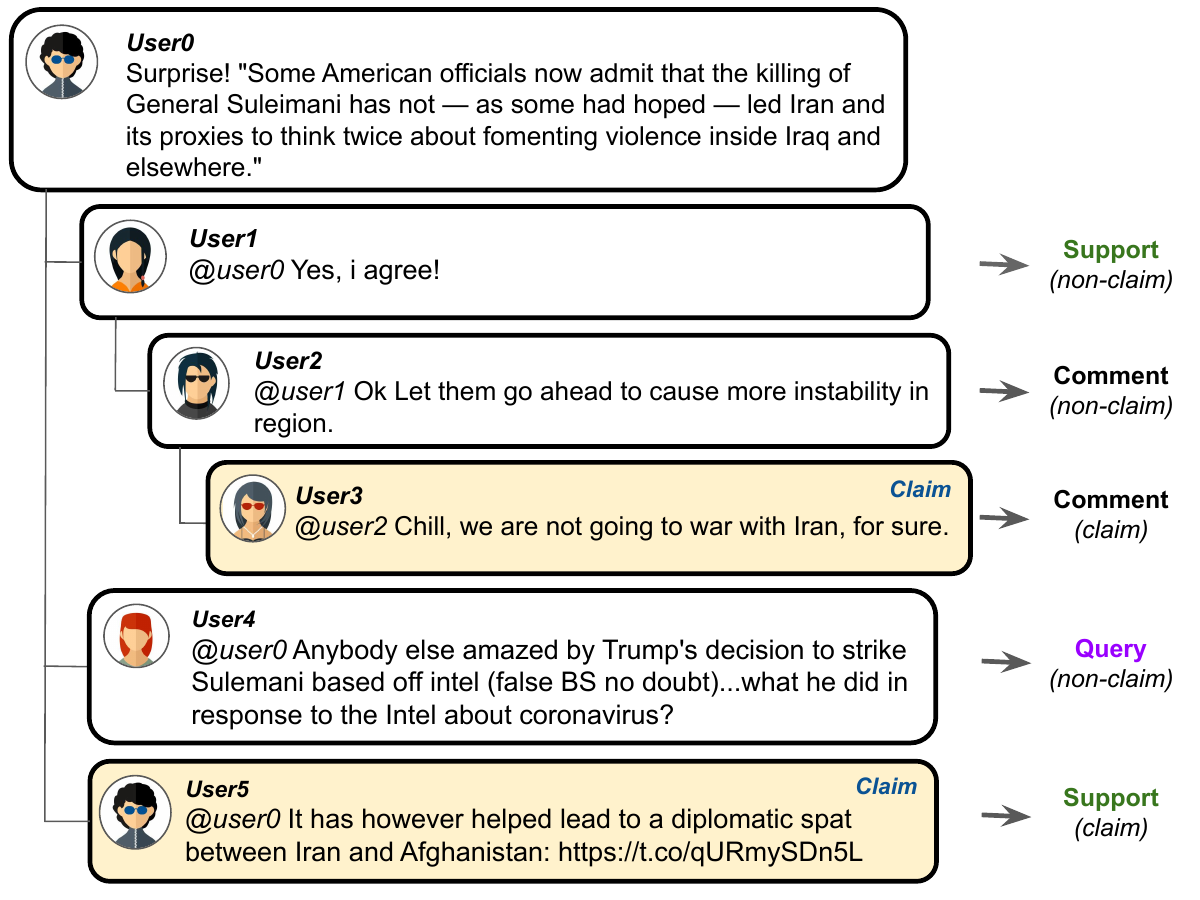}}
\hfill
\subfloat[\footnotesize{Misinformation}]{\includegraphics[width=0.45\textwidth]{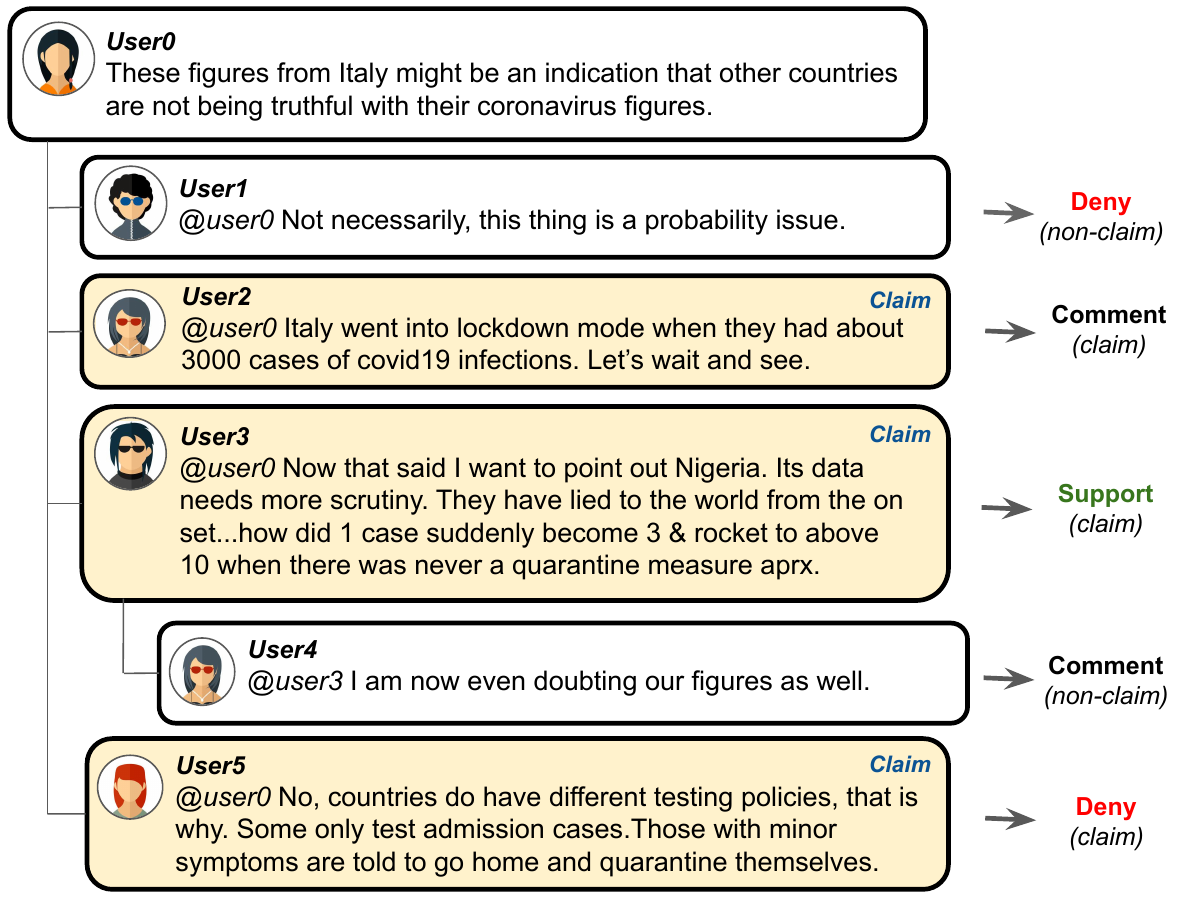}}

\caption{Illustrative examples of Twitter posts and their subsequent conversational thread, with the original post containing false and true information, respectively. The hierarchical arrangement of replies reflects users' interaction and stances toward the source post. Replies highlighted in yellow denote claims made by the users in response to the source post.}
\label{fig:motivation}

\vspace{-1em}
\end{figure*}

\textbf{\textit{Crowd intelligence.}}
Social media users often express their opinions on posts by supporting or contradicting them, which aids in determining the credibility of the information posted \cite{zubiaga2016analysing, yang2024reinforcement}. Building on this, we introduce the concept of \textit{crowd intelligence}. This refers to the collective decision-making capabilities of social media users that emerge from the interaction and aggregation of diverse inputs. Researchers highlight that stances such as support and denial play a crucial role in predicting veracity, whereas comments -- statements lacking any explicit stance --  are deemed irrelevant for this task \cite{ma2024dsmm}. However, we hypothesize that not all comments are irrelevant, and they may contain crucial claims essential for assessing the truthfulness of the information posted. As defined by Toulmin \cite{toulmin2003uses},  \textit{a claim is an assertive statement with or without evidence}. These assertive statements in user responses can be strong indicators for determining the veracity of the source post, giving additional context for the post. For instance, in Figure \ref{fig:motivation}(a),
where \textbf{User3}'s comment lacks an explicit stance but makes an important claim that cannot be overlooked. Thus, by harnessing crowd intelligence through user stances and claims, we can significantly enhance the effectiveness of misinformation prediction. Figure \ref{fig:motivation} vividly demonstrates the utility of analyzing the source posts and how others orient to them. It depicts two posts from Twitter and some initial user replies. The first example is a tweet about the aftermath of General Suleimani's death and its ineffectiveness in reducing violence from Iran and its proxies. Importantly, this tweet is not misinformation.
The reactions are mostly favorable or add additional commentary, indicating a shared understanding of the situation described in the tweet. This shared viewpoint implies a fundamental level of public acceptance, which is critical in determining the veracity of information shared on social media at an early stage. Furthermore, claims in this conversation thread elevate the conversation above mere agreement. Responses like \textit{`we are not going to war with Iran, for sure'} are significant as they go beyond straightforward agreement to make independent claims reinforcing the original post. These claims are critical; they carry an inherent legitimacy as an authority contrasted with passive responses such as \textit{`Yes, I agree.'} These claims serve as anchors in the discourse, providing substantive support for the tweet's message while highlighting a higher level of community engagement. In contrast, in the second case, the tweet is misinformation.
The discourse surrounding this tweet is characterized by a broad spectrum of stances, most notably the presence of direct denials and contradictory claims that serve as significant red flags here. These claims directly contradict the tweet's assertions, showing denial of the source post. This exemplifies that crowd intelligence is critical in determining social media content's credibility.

To address the early misinformation prediction problem, we propose \name, a deep learning framework that utilizes the post’s
semantic and user responses to predict false information. At a high level, \name\ first adopts a Q-network to learn propagation representation through the user's stances towards the source post. Next, the model utilizes a transformer-based model to learn the semantic features from the source post and the relevant replies. Finally, the learned semantic and propagation representations are combined to determine whether the post is misinformation or not. We also create a real-world corpus, \dataset, comprising 199 Twitter posts followed by 14,436 user replies. Each source post is manually annotated for misinformation, and user replies are labeled with claim and stance labels.

\textbf{\textit{Our contributions.}} Through this work, we make the following contributions:
\begin{itemize}[leftmargin=1em, itemsep=0pt]
    \item We present a novel model for early misinformation detection that takes advantage of crowd intelligence reflected through user reply stances and claims.    
   
    \item We present a novel dataset that includes Twitter threads meticulously annotated for the veracity of the source post. Additionally, we mark stances expressed in replies to the source post and claim labels associated with each reply. 

    \item We perform exhaustive experiments and confirm significant improvements over several baselines. We also provide detailed qualitative and quantitative analysis.

\end{itemize}

\section{Related Work}
The widespread dissemination of false or misleading information on social media platforms has attracted much scrutiny in recent years. A plethora of research has been conducted focusing on various aspects of identifying and mitigating misinformation, including fake news, rumors, etc \cite{zhou2020survey, ye2024tai, jain2024confake, choudhary2021linguistic}. Researchers have investigated several approaches for identifying misinformation, including methods that analyze the content, examine the network, and predict falsehoods in advance.

\textbf{\textit{Content-based detection.}} Content-based approaches rely heavily on linguistic clues such as writing styles, lexical aspects, sentiment analysis, and subject relevance \cite{ye2024tai, jain2024confake, choudhary2021linguistic}. For example, Castillo et al. \cite{castillo2011information} discovered that highly reputable social media messages have more URLs and longer text lengths than less credible ones. Similarly, Pawan et al. \cite{verma2021welfake} introduced the WELFake model, which detects bogus news by combining word embedding vector and linguistic data. Anshika et al. \cite{choudhary2021linguistic} proposed a deep learning model that leverages syntactic, grammatical, semantic, and readability aspects of news stories to identify bogus news. Validated over several publicly accessible datasets, Jain et al. \cite{jain2024confake} created the Confake algorithm using a comprehensive set of content-based features and word vector attributes taken from news items. These content-based methods analyze the credibility of individual posts in isolation, often neglecting the high correlation between a post's veracity and the replies it garners from other users.

\textbf{\textit{Propagation-based detection.}} To overcome the limitations of content-based methods, current research has concentrated on propagation-based approaches. These methods employ social context knowledge to identify false information by examining the dissemination of information on social networks, the people responsible for disseminating it, and the relationships between these propagators \cite{ma2018rumor, dhawan2024game, song2021temporally, wu2023adversarial}. Ma et al. \cite{ma2018rumor}  developed a neural network with a tree structure that uses false news cascades to detect disinformation. Dhawan et al. \cite{dhawan2024game} introduced GAME-ON, a system that utilizes Graph Neural Networks to enable detailed interactions inside and between several modalities for the purpose of detecting multimodal fake news. Song et al. \cite{song2021temporally} created a temporal learning model based on graphs to capture the changing patterns of tweets that are organized as tree-structured data. Kang et al. \cite{kang2021fake}  constructed a news detection graph that links bogus news with many third-party information sources. Sun et al. \cite{sun2022structure} proposed a hypergraph learning model that utilizes a unique hyperedge walking technique and hyperedge expansion method to create comprehensive representations for entire graphs. These studies indicate that models utilizing network information outperform those that merely rely on content. In addition to propagation features, user-specific attributes, including the number of followers and the stance of other users toward the post, have also been employed to detect misinformation. The significance of other user's reactions in the verification of information has been emphasized in numerous studies \cite{castillo2011information, liu2015real, enayet2017niletmrg}. Li et al. \cite{li2016user} conducted a study on the semantic aspects of false information, analyzing their dissemination and user characteristics, using 421 false statements and 1.47 million related tweets. One of the most significant discoveries is that individuals are inclined to disseminate falsehoods when they are uncertain about their veracity without incorporating their personal opinions.  According to Zubiaga et al. \cite{li2016user}, users are inclined to endorse unverified information. In contrast, highly reputable users, such as news organizations, endeavor to publish well-reasoned statements that are supported by evidence and appear to be certain. These findings underscore that models using network information and user responses are highly efficient for misinformation detection on social media.

\textbf{\textit{Early detection of misinformation.}} With the quick dissemination of information on social media, false information can reach thousands of users in minutes, causing significant confusion, panic, and even physical harm. It is essential for real-time systems to identify misinformation at its inception before its rampant dissemination. Early detection algorithms have been the subject of numerous investigations \cite{kwon2017rumor, wu2024graph, liu2018early}. Kwon et al. \cite{kwon2017rumor} conducted an analysis of feature stability over time. They discovered that user and linguistic features are more effective than structured and propagation features in determining the veracity of information during the early stages. Using network embedding techniques on social network graphs, Liu and Wu  \cite{liu2018early} constructed user representations using network embedding approaches on the social network graph. The critical importance of this area in real systems necessitates additional research despite the existence of current research.

An effective solution to address the issue of misinformation involves a detailed exploration of user responses to posted information, considering their stances and the strength of their claims. The positions held by users on the information can offer valuable insights into the accuracy of a certain post. The combined intelligence of users can help identify inconsistencies or provide support for factual content. Traditional network or user-based approaches often neglect these factors, instead primarily concentrating on disparities in structure. Addressing these limitations, we propose a framework that integrates linguistic and crowd intelligence to enhance the integrity and reliability of online information by swiftly identifying misinformation. Our framework is designed to meet the critical need for early detection, providing a robust solution to mitigate the rapid spread of misinformation at the earliest possible.

\section{Dataset}
Over the past few years, several misinformation detection datasets have been released \cite{zubiaga2016analysing, gorrell2019semeval}. However, none of these datasets come with both stance and claim labels for the user replies. The only dataset that closely aligns with our research objectives is the SemEval-2019 Task 7 dataset \cite{gorrell2019semeval}. It contains 446 conversation threads and 8142 user replies, containing rumors gathered from Twitter and Reddit during eight distinct events. 

It contains stance labels (i.e., \textit{support}, \textit{deny}, \textit{comments}, and \textit{query}) for replies and veracity labels (i.e., \textit{true}, \textit{false}, and \textit{unverified}) for source posts. However, similar to other cases, it also does not contain claim labels. For our task, we annotate claim labels at the user reply level. 
Given the age of these events, we further meticulously cross-checked all source posts for veracity labels to ensure the accuracy and reliability of the data. Furthermore, we revisit all instances labeled as \textit{unverified} to reassess their truthfulness with new evidence. Out of 110 unverified tweets, 47 have been classified as misinformation, while 63 as non-misinformation. Similarly, 8 out of the 13 unverified Reddit posts are categorized as misinformation and 5 as true.

\subsection{The \dataset\ dataset}
Despite its usefulness, the SemEval-2019 dataset has some limitations -- it primarily focuses on well-known and limited topics. Also, the dataset contains posts related to events that are now outdated and of limited relevance. 
Therefore, we curate a new Twitter corpus, \dataset\ (\textbf{Mis}information detection on \textbf{T}witter) with a much diverse set of recent topics, hence making it more representative of general social media interactions.

\paragraph*{\textbf{Data collection}}
We collect random trending Twitter threads from various periods, user demographics, and thematic categories. This ensures that the dataset is representative of a cross-section of Twitter discourse and encompasses various interests and perspectives. We leverage Twitter's API to access publicly available conversation threads while adhering to ethical guidelines and platform policies regarding data usage and user privacy.\footnote{\url{https://x.com/en/privacy}} We collect the data between October 4, 2022 to December 31, 2022. Each instance selected for inclusion consists of a source post, which serves as the initiating message or topic of discussion, followed by a series of user replies, forming a cohesive conversational thread. 

\paragraph*{\textbf{Data annotation}}
For annotation, we employ four annotators experienced in social media and linguistics. We annotate the collected samples focusing on two dimensions -- \textit{source post} and \textit{user replies}.

\begin{itemize}
\item {Source Post:} We meticulously examine every source post and annotate it to indicate whether it contains misinformation or adheres to factual accuracy. For each tweet, the annotators are free to assess the content's veracity based on their prior knowledge or available external sources.  

\item {User Replies:} We analyze each reply of the conversation thread and annotate them for their stance towards the source post. Stances can take one of the four values: support, deny, comment, and query. Note that we use \textit{`root'} to denote the stance of the source post. Besides stance annotation, user replies are subject to claim annotation to identify whether they assert something or not. We employ the annotation guidelines provided by Gupta et al. \cite{gupta2021lesa} to annotate the replies for claims. 
\end{itemize}

To ensure consistency and agreement among annotators, the annotation process involves multiple rounds of review and refinement. Any discrepancies were resolved through consensus discussions and adjudication by senior annotators or domain experts. This iterative annotation process aimed to achieve high-quality annotations with a high degree of accuracy, reliability, and relevance to the task of misinformation detection and analysis on social media platforms. To support our claim, we conduct inter-rater reliability and compute Cohen's Kappa \cite{cohen1960coefficient} inter-annotator agreement (IAA) score as 0.67. \footnote
{We present extended annotation guidelines in {\textcolor{blue}{Supplementary}}.} 

\subsection{Dataset Statistics and Analysis}
\dataset\ comprises 199 tweets with 14,436 user replies. 
Table \ref{tab:dataset-stats} contains detailed statistics for both \dataset\ and SemEval datasets. It is important to note that the SemEval dataset has a higher prevalence of misinformation than ours, while the number of misinformation instances in \dataset\ is significantly low. This is due to the fact that our dataset is \textit{neutrally-seed} and does not focus on specific and controversial events, such as those highlighted in the SemEval dataset. The qualitative analysis of the dataset is delineated as follows.

\begin{table}[ht]
\centering
\caption{Dataset statistics showing misinformation (mis) and non-misinformation (non-mis) labels. \label{tab:dataset-stats}}
\begin{tabular}{ccccccc}
\toprule
\textbf{} & \multicolumn{2}{c}{\textbf{\dataset}} & \multicolumn{2}{c}{\textbf{SemEval$_{Twitter}$}} & \multicolumn{2}{c}{\textbf{SemEval$_{Reddit}$}} \\ \cmidrule{2-7} 
\textbf{} & \textbf{Mis}  & \textbf{Non-mis}  & \textbf{Mis} & \textbf{Non-mis} & \textbf{Mis} & \textbf{Non-mis}   \\ \midrule
Train     & 40              & 119             & 190                 & 135                  & 12                  & 28                  \\
Test      & 10              & 30              & 24                  & 32                   & 14                  & 11                  \\ \midrule
Total     & 50              & 149             & 214                 & 167                  & 26                  & 39                  \\ \bottomrule
\end{tabular}
\end{table}

\paragraph*{\textbf{Stance evolution within replies}}

\begin{figure*}[!ht]
\centering
\subfloat[\footnotesize{Misinformation}]{\includegraphics[width=0.49\textwidth]{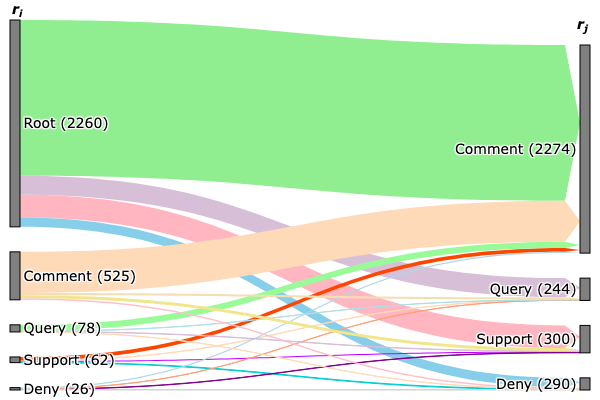}}
\hfil
\subfloat[\footnotesize{Non-misinformation}]{\includegraphics[width=0.49\textwidth]{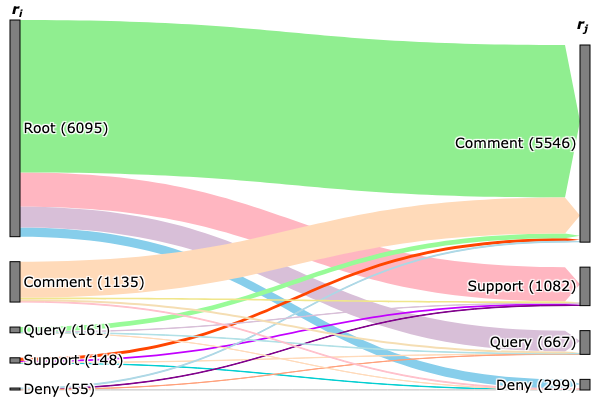}}
\caption{Analysis of stance evolution in replies within the conversation threads. The diagram depicts the change of stance from reply $r_i$ (shown on the left vertical axis) to subsequent reply $r_j$ (shown on the right vertical axis) in conversation threads.}
\label{fig:stance_misinformation}
\end{figure*}

We analyze how stances change between consecutive responses within the conversation threads. Figure \ref{fig:stance_misinformation} demonstrates the evolution of stance from reply $r_i$ (left vertical) to subsequent reply $r_j$ (right vertical) in conversation threads. The first vertical represents the stance of a reply, with the caveat that stances for source posts are labeled as `root' due to their inherent nature. Conversely, the second vertical denotes the stance of the immediate following reply. Upon analysis, we observe a notable trend across both misinformation and true information -- most direct replies to source posts are categorized as `comments,' lacking explicit stances, thus making it crucial to analyze further discourse to understand the veracity of the social media threads. However, a distinction emerges regarding the level of support for the source post between true and false misinformation. Specifically, in the case of misinformation, the support for the source post is notably lower than true information, as evidenced by the flow from `root' to `support' in Figure \ref{fig:stance_misinformation}(b). 

\paragraph*{\textbf{Relationship between reply stance and claims}}

\begin{figure}[!ht]
\centering
\includegraphics[width=0.45\textwidth]{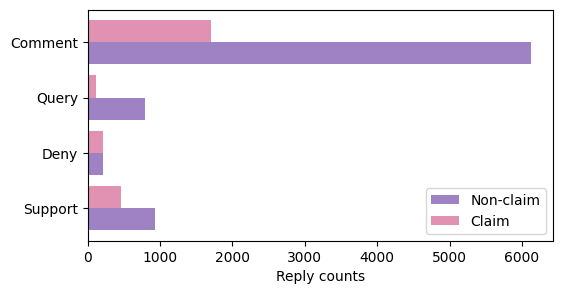}
\caption{Correlation between different stances towards source posts and their assertive nature, as indicated by claim labels, across all conversation threads. The purple bar indicates whether the response is a non-claim, whereas the pink bar indicates whether it is a claim. \label{fig:stance_claim}}
\end{figure}

In this analysis, we delve deep into the relationship between reply stances and the presence of claims within them. We discover intriguing patterns that shed light on user behavior and communication dynamics. Figure \ref{fig:stance_claim} depicts the key findings from this analysis. Initially, we observe a significant predominance of the `comment' stance in responses without explicit claims. This implies that users frequently engage in discussions, share opinions, and provide context without making factual claims. However, in replies containing claims, the prevalence of the `comment' stance reduces to more than half compared to non-claims. However, the claim count remains higher when compared to other stances, indicating that not all explicit stances are useful and rejecting all comments can be harmful. The `deny' stance is notably uncommon across all responses, whether they contain claims or not. This suggests that users prefer to present alternative viewpoints or evidence rather than reject statements outright. However, it is worth noting claims and non-claims appear equally within the `deny' stance, emphasizing its importance as a stance to consider when analyzing user responses.

\section{Methodology}
\subsection{Problem Statement} 

In this work, we define misinformation as \textit{`a piece of information propagating through social media which is ultimately verified as false or inaccurate.'} Formally, we define the task as follows. 
    Given a social media post $p$ and its associated set of replies $R = \{ r_1, r_2, \dots, r_n \}$. Each reply $r_i$ is labeled with stance label $s_i$ (i.e., support, deny, query, or comment) towards $p$ and claim $c_i$ (i.e., claim or non-claim). Our objective is to predict whether $p$ disseminates misinformation as early as possible. This involves analyzing the initial few replies, denoted as $\tau$, and leveraging the stance $s_i$ and claim $c_i$ labels associated with these replies.

\subsection{Proposed Framework}
\begin{figure*}[!th]
\centering
\includegraphics[width=\textwidth]{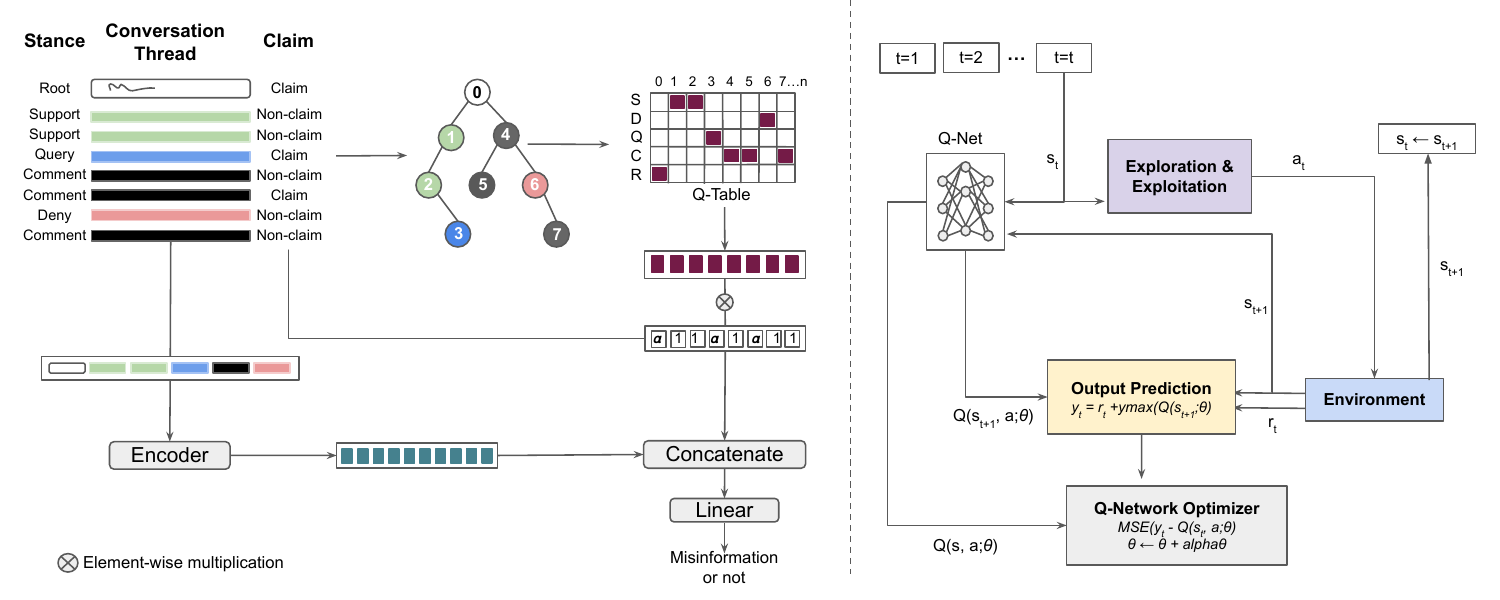}
\caption{Illustrative model diagram for our proposed framework for early misinformation prediction. The right side of the diagram shows the Q-table update mechanism. S, D, Q, and C in Q-table denote support, deny, query, and comment, respectively.
}
\label{fig:model}
\end{figure*}

In our work, we propose a novel framework, \name, to comprehend the social media conversation thread propagation dynamics using deep Q-learning, an off-policy reinforcement learning technique. Our primary objective is to understand the stances observed within the replies of a thread. We opt for deep Q-learning due to its capability to effectively navigate complex decision spaces and learn network properties well, thereby shedding light on the underlying discourse patterns. By leveraging deep Q-learning, our model can iteratively explore the state-action space, gradually refining its understanding of the underlying thread propagation dynamics. Furthermore, the off-policy nature of deep Q-learning enables the decoupling of the exploration and exploitation phases, facilitating more robust and efficient learning. Figure \ref{fig:model} illustrates the overall architecture of \name.

To begin with, we define the foundational elements of our methodology:

\textbf{\textit{State Space:}} The \textit{state} is defined as the content and context of the current reply. Thus, the state space is determined by the set of all possible replies and their content. 
Mathematically, a state $s$ can be represented as,
\begin{equation}
    s = (i,c,p)
\end{equation}
where $i$ denotes the reply id, $c$ represents the content of the current reply, $p$ denotes time of posting. $s_0$ represents the source post in the conversation thread. 
Consequently, the state space (S) is represented as follows,
\begin{equation}
    S = \{(i_0,c_0,p_0), (i_1,c_1,p_1), (i_2,c_2,p_2),...(i_n,c_n,p_n)\}
\end{equation}
where each tuple $(i_t,c_t,p_t)$ represents a specific state in the state space and $n$ denotes the total number of replies (states) in the conversation thread.  

\textbf{\textit{Action Space:}} Complementing the state space, the action space plays a pivotal role in orchestrating the traversal of the conversation thread. Our architecture conceptualizes the stances expressed within the replies as \textit{action}. Formally, at time step, $t$, the action $a_t$ signifies the stance of the corresponding reply within the thread towards the source post. Thus, the action space encompasses all possible stances a reply can exhibit. Notably, the source post, the initial point of the conversation thread, is characterized by the \textit{Root} stance. 
Consequently, the action space (A) is delineated as follows,
\begin{equation}
    A = \{support, deny, query, comment, root\}
\end{equation}
with $a_0$ representing the state of the source post, always being assigned the \textit{root} stance.

\subsection{Q-Learning Phase}
In the Q-learning phase, the primary objective is to accurately predict the Q-values for each state-action pair within the model's architecture. In essence, these Q-values represent the expected rewards of having a particular state at the given action. We employ a deep Q-learning network, an effective hybrid neural network framework, which would take in the current state of the environment as input and output the estimated Q-values for each possible action. The learning process begins with initializing the Q-table entries to zero. This ensures that the learning starts from a neutral and unbiased perspective. As the model interacts with the environment, it transitions from one state to another based on the actions taken, and it receives rewards that reflect the efficacy of each action. Learning directly from consecutive samples is inefficient due to the strong correlations between the samples and the chances of overfitting; thus, randomizing the samples limits these correlations and reduces the variance of the updates. We leverage exploration and exploitation within the framework through $\epsilon$-greedy policy to solve this. This policy allows the agent to explore new actions with a certain probability, determined by $\epsilon$. The value of $\epsilon$ ranges between 0 and 1. Exploration involves randomly selecting actions, while exploitation entails choosing the action with the maximum Q-value obtained from the Q-network table. The Q-network table is updated using the Bellman equation:
\begin{equation}
    Q(s_t, a_t) = R_t + \epsilon \max_a Q(s_{t+1}, a)
\end{equation}
where $Q(s_t, a_t)$ is the Q-value for the state-action pair at time step $t$, $r_t$ is the immediate reward obtained at that state, $\epsilon$ is the discount factor that balances the importance of immediate and future rewards, and $s_{t+1}$ is the next state after taking action. The agent learns to make decisions that optimize long-term rewards by maximizing the Q-value for each state-action pair. We define reward function $R$ as the sum of the claim reward $R_{\text{claim}}$ and the stance reward $R_{\text{stance}}$. The claim reward $R_{\text{claim}}$ is 1 if the content of the reply is a claim, denoted by $c$, and 0 otherwise. The stance reward $R_{\text{stance}}$ is defined based on the stance taken in the reply. Specifically, we formally define $R_{\text{claim}}$ and $R_{\text{stance}}$ as,

\begin{equation}
    R_{\text{claim}} =
    \begin{cases}
    1 & \text{if } c \text{ is a claim} \\
    0 & \text{otherwise}
    \end{cases}
\end{equation}

\begin{equation}
    R_{\text{stance}} = 
    \begin{cases}
    1 & \text{if } stance = \text{ `support' or `query' or `root'} \\
    -1 & \text{if } stance = \text{ `deny'} \\
    0 & \text{if } stance = \text{ `comment'}
    \end{cases}
\end{equation}
Therefore, the total reward $R$ ranges between -1 to 2 and is given by,
\begin{equation}
    R = R_{\text{claim}} + R_{\text{stance}}
\end{equation}

To optimize the parameters of the Q-network, we use gradient descent with the Mean Squared Error loss. The target value $y_t$ is calculated as follows,
\begin{gather}
    y_t = \mathbb{E}_{s'}\left\lbrace R + \epsilon \max_{a} Q(s_{t+1}, a, w_{t-1})\right\rbrace 
\end{gather}
The loss function $L_t(w_t)$ is then defined as,
\begin{gather}
    L_t(w_t) = \left[y_t - Q(s, a, w_t) \right]^2  
\end{gather}
where $Q(s, a, w_t)$ is the predicted Q-value for state $s$ and action $a$ with current parameters $w_t$. 


By minimizing this loss using gradient descent, the parameters of the Q-network are updated to approximate the true Q-values better. We employ the Adam optimizer during backpropagation to efficiently update the parameters and improve the overall performance of the Q-network.

\subsection{Training Phase}
After the Q-learning phase, we obtain a Q-table. The values in the Q-table represent the expected future rewards for taking a particular action in a given state. To construct a feature vector for each conversation thread, we extract the values from the Q-table. This involves capturing the Q-values corresponding to the given stance exhibited within the thread for each reply. We define the feature vector for thread $i$ as follows,

\begin{equation}
    F_i = [Q(s_0, a_0), Q(s_1, a_1),\cdots , Q(s_n, a_n)]
\end{equation}
where $Q(s_j, a_j)$ denotes the Q-value for $jth$ reply in the thread, ($s_j$, $a_j$) denotes its state-action pair for the reply, and $n$ is the total number of replies in the thread. 

For each reply in the thread, we create a binary claim vector $C_i$, of the same size as the thread. If the reply is a claim, the corresponding element in $c_{ij}$ $\in$ $C_i$ is marked as 1 else 0. So, the claim vector for the $ith$ thread would be:

\begin{equation}
    C_i = [c_{i0}, c_{i1}, c_{i2}, \cdots, c_{in}]
\end{equation}
where 
\begin{equation*}
    c_{ij} = 
    \begin{cases}
    1 & \text{if $jth$ reply in $ith$ thread is a claim} \\
    0 & \text{otherwise}
    \end{cases}
\end{equation*}

With the objective of assigning more weightage to claims in the veracity determination process, we first multiply each value in the claim vector $C_i$ by the scalar claim weight factor $\alpha$. Then, we compute the Hadamard product of the resulting vector with the feature vector $F_i$, 

\begin{equation}
    F_i =( \alpha \bullet C_i) \otimes F_i
\end{equation}
where $\alpha$ = 1 denotes equal emphasis on claims and non-claims in the reply thread. The value of  $\alpha$ can be empirically set to an appropriate real number.

Next, we combine the source post and its corresponding replies having \textit{support}, \textit{deny}, and \textit{query} stances and obtain rich semantic representation $S$ for the combined text. We utilize a pretrained language model -- BERT \cite{devlin-etal-2019-bert} -- and fine-tune it on our data to generate text representation.
\begin{equation}
    S_i = BERT(combined\_text_i)
\end{equation}

Subsequently, we concatenate Q-feature vector $F_i$ with the semantic feature vector $S_i$ to form the final feature vector $V_i$ for each thread, 

\begin{equation}
    V_i = [F_i, S_i]
\end{equation}

Finally, we employ a linear layer to facilitate final classification that can be computed as, 
\begin{equation}
    \hat{y} = \sigma(\textbf{W}V + b)
\end{equation}
where $\hat{y}$ is the final predicted misinformation label. \textbf{W} represents the weight matrix and $b$ represents the bias vector of the linear layer.

\section{Experimental Settings}
\subsection{Baseline Models}

We employ the following baseline systems. $\triangleright$ \textbf{LSTM} \cite{hochreiter1997long}: A long short-term memory network trained for binary classification tasks. $\triangleright$ \textbf{RNN} \cite{grossberg2013recurrent}: A recurrent neural network is any network whose neurons send feedback signals to each other. $\triangleright$ \textbf{BERT} \cite{devlin-etal-2019-bert}: A bidirectional transformer-inspired auto-encoder language model fine-tune for our misinformation detection task. $\triangleright$ \textbf{RoBERTa} \cite{liu2019roberta}: A robustly optimized BERT approach with improved training methodology. We fine-tune it on our data. $\triangleright$ \textbf{EventAI} \cite{eventai2019semeval}: The system achieved the highest ranking in the SemEval 2019 Task 7 on rumor verification. To verify rumors, they utilized information from several aspects, including the content of the rumor, the reliability of the source, the credibility of the user, and the stance of the user. We exclude the user credibility features due to the absence of relevant user information. $\triangleright$ \textbf{GCN} \cite{bacciu2020gentle}: A graph convolutional network is a neural network adapted to leverage the structure and properties of graphs. We also augmented it with word embeddings. $\triangleright$ \textbf{RPL} \cite{lin2023zero}: A zero-shot framework based on a prompt learning system to detect rumors falling in different domains. It uses a hierarchical prompt encoding mechanism for learning language-agnostic contextual representations for both prompts and data. $\triangleright$ \textbf{ACLR} \cite{lin2022detect}: The architecture proposes an adversarial contrastive learning framework to detect rumors. $\triangleright$ \textbf{GMVCN} \cite{wu2024graph}: The system encodes the multiple views of the conversation thread based on GCN and leverages convolutional neural networks to capture consistent and complementary information.  $\triangleright$ \textbf{GPT-4} \cite{achiam2023gpt}: A state-of-the-art large language model intended to address intricate queries and provide responses that are both coherent and contextually pertinent across a diverse array of tasks.

\subsection{Experimental Setup and Evaluation Metrics}
In the Q-training phase, we use a single-layer network for the deep Q-learning network and train it over $1000$ episodes. To balance exploration and exploitation, we empirically set the discount factor ($\epsilon$) at $0.2$, resulting in less exploration during training. On our dataset, we fine-tune a BERT model obtained from HuggingFace.\footnote{\url{https://huggingface.co/docs/transformers/en/model_doc/bert}} The fine-tuned BERT model is then used to generate text embeddings for the final feature vectors. We train this model for $20$ epochs, stopping early based on macro-F1 and setting patience to $3$. The final classifier employs a feed-forward layer with softmax activation. We use the Adam Optimizer with a learning rate of 0.001 and a batch size of 8 for training. Experimentally, we find the best $\alpha$ value to be 2 (shown in Table \ref{tab:ablation-alpha}). We keep a development set containing $10\%$ of the training data to optimize and fine-tune the models, allowing us to monitor and adjust performance during training iterations. For all baseline systems, we replicate the models using the descriptions provided in their original papers. For evaluation, we primarily use macro-F1 scores from \textit{scikit-learn} library. In addition, we report class-wise recall, precision, and F1 scores to provide a more detailed understanding of model performance.

\section{Results and Analysis}
\begin{table*}[ht]
\centering
\caption{Experimental results of \name\ and its variants (last two rows) on our dataset  \dataset.  Input to the system: \textit{SP} denotes Source Post only, and \textit{SP $\cup$ CT} denotes Source Post and the corresponding Conversation Thread. \label{tab:results}}

\begin{tabular}{lcccccccc}
\toprule
\multicolumn{1}{c}{\multirow{2}{*}{\textbf{Model}}} & \multicolumn{1}{c}{\multirow{2}{*}{\textbf{Input}}} & \multicolumn{3}{c}{\textbf{Non-misinformation}}       & \multicolumn{3}{c}{\textbf{Misinformation}}                 & \multirow{2}{*}{\textbf{Macro-F1}} \\ \cmidrule{3-8}

\multicolumn{1}{c}{} & \multicolumn{1}{c}{} & \textbf{Pecision} & \textbf{Recall} & \textbf{F1} & \textbf{Pecision} & \textbf{Recall} & \textbf{F1} &                                    \\ \midrule
\textcolor{black}{LSTM}     & \textit{SP}  & 0.7838 & 0.9667 & 0.8657 & 0.6667 & 0.2000    & 0.3077 & 0.5867 \\
\textcolor{black}{RNN}    & \textit{SP}          & 0.7586 & 0.7333 & 0.7458 & 0.2727 & 0.3000    & 0.2857 & 0.5157 \\
\textcolor{black}{BERT}     & \textit{SP}        & 0.8000    & 0.8000    & 0.8000    & 0.4000    & 0.4000    & 0.4000    & 0.6000    \\
\textcolor{black}{RoBERTa}  &   \textit{SP}  &  0.7812 & 0.8333 & 0.8065 & 0.3750  & 0.3000    & 0.3333 & 0.5699 \\

\midrule
BERT$_{\textit{stance}}$     &  \textit{SP $\cup$ CT} & 0.7188    & 0.7667 & 0.7419 & 0.1250 & 0.1000 & 0.1111 & 0.4265 \\
BERT$_{\textit{claim}}$      & \textit{SP $\cup$ CT} &  0.7692     & 1.0000 & 0.8696 & 1.0000     & 0.1000 & 0.1818 & 0.5257 \\
BERT$_{\textit{stance\_claim}}$  & \textit{SP $\cup$ CT} &  0.7429 & 0.8667 & 0.8000    & 0.2000   & 0.1000 & 0.1333 & 0.4667 \\ 
\midrule

EventAI    & \textit{SP $\cup$ CT}  & 0.7500   & 1.0000  &   0.8571 &  0.0000  &  0.0000  &  0.0000 & 0.4286 \\
GCN    & \textit{SP $\cup$ CT}   & 0.8000    & 0.6667 & 0.7273 & 0.3333 & 0.5000    & 0.4000    & 0.5636 \\ 
\textcolor{black}{ACLR}   & \textit{SP $\cup$ CT}  & 0.7742 & 0.9600   & 0.8571 & 0.6667 & 0.2222 & 0.3333 & 0.5952 \\
\textcolor{black}{RPL}   & \textit{SP  $\cup$ CT}    & 0.7500   & 0.7500   & 0.7500   & 0.3000    & 0.3000    & 0.3000    & 0.5250  \\
\textcolor{black}{GMVCN}     & \textit{SP  $\cup$ CT}    & 0.6957 & 0.5333 & 0.6038 & 0.1765 & 0.3000    & 0.2222 & 0.4130  \\

\midrule
GPT-4    &  SP & 0.7619 & 0.5333 & 0.6275 & 0.2632 & 0.5000 & 0.3448 & 0.4861 \\
GPT-4 & \textit{SP  $\cup$ CT} & 0.7667 & 0.7667 & 0.7667 & 0.3000    & 0.3000 & 0.3000   & 0.5333 \\
\midrule

\rowcolor[rgb]{0.839,0.882,1} \name    & \textit{SP $\cup$ CT}  & 0.8125 & 0.8667 & 0.8387 & 0.5000    & 0.4000    & 0.4444 & 0.6416
\\

\midrule
$\quad-$ \{\textit{Q-learning}\} &  \textit{SP $\cup$ CT}  & 0.7879 & 0.8667 & 0.8254 & 0.4286 & 0.3000 & 0.3529 & 0.5892 \\
$\quad-$ \{\textit{Text Features}\} &  \textit{SP $\cup$ CT}  & 0.7576 & 0.8333 & 0.7937 & 0.2857 & 0.2000 & 0.2353 & 0.5145 \\

 \bottomrule
\end{tabular}
\end{table*}

Table \ref{tab:results} presents our consolidated results on our dataset \dataset.\footnote{See \textcolor{blue}{Supplementary} for results on SemEval$_{Twitter}$ and SemEval$_{Reddit}$ datasets.} We ran two sets of experiments to evaluate text classification models. Initially, models were trained solely on the source post (SP), with no additional information from conversation threads (CT). Second, we incorporated baseline systems that utilized conversation threads into the misinformation identification process, employing a variety of methodologies to capture response dynamics. The first two rows of Table \ref{tab:results} show the performance of binary classification systems for misinformation detection with only the post. Notably, BERT outperformed other systems on the Twitter dataset, with the highest macro-F1 score of $0.60$. To improve the source post, we added replies tagged as `support,' `deny,' or `query,' as well as claims, to the BERT model. Among these configurations, the one that included only replies classified as claims alongside the source post produced the best results, with a macro-F1 score of $0.52$ macro-F1 (rows 5-7). This emphasizes the significance of claims in conversation threads as indicators of misinformation. Interestingly, including these replies did not significantly improve the overall BERT performance compared to using the source post alone, highlighting the challenge posed by the potential noise in the replies. Moreover, compared to the existing baseline systems that include the conversation threads in their frameworks, \name\ achieved the highest macro-F1 score, exhibiting approximately $5\%$ improvement over the best baseline -- ACLR \cite{lin2022detect}. Furthermore, \name\ obtained an F1 score of $0.44$ for the true class, indicating superior performance in detecting misinformation compared to the other systems. GPT-4 initially performed poorly for the task, scoring $0.48$. However, when replies were included, its performance improved significantly, with a nearly $4.7\%$ increase. This improvement could be attributed to the context provided by the conversation threads. By taking into account interactions within the thread, GPT-4 was likely able to understand the conversation's context and nuances better, resulting in improved performance in identifying and addressing misinformation.

\subsection{Early Detection Efficiency}

\begin{figure}[!ht]
\centering
\includegraphics[width=0.48\textwidth]{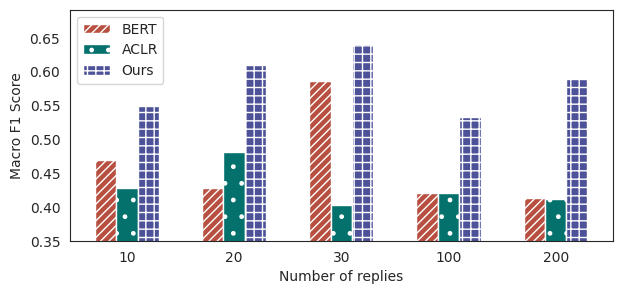}
\caption{Macro-F1 scores are presented for our model \name\ (indicated by the violet bar with vertical lines) compared to the top baseline systems include conversation threads: ACLR (represented by the green bar with dots) and BERT (depicted by the red bar with diagonal lines). The evaluation was conducted across varying numbers of replies within the conversation thread.
}
\label{fig:early_detection}
\end{figure}

Detecting misinformation early is crucial to mitigate its societal impact. We establish detection milestones, such as the number of reply posts, to identify content for evaluation only up to these points. We progressively analyze test data chronologically until reaching the desired number of posts. Figure \ref{fig:early_detection} illustrates the effectiveness of our approach \name, compared to two best-performing baselines that use conversation threads -- ACLR \cite{lin2022detect} and BERT$_{\textit{claim}}$ (denoted as BERT) for early detection. For fewer replies (10, 20, and 30), \name\ consistently outperforms both BERT and ACLR. Our Q-learning-based method consistently outperforms other approaches from the beginning, demonstrating its effectiveness in the early detection of misinformation. Our model achieves a high macro-F1 score shortly after its initial dissemination. Using only the first ten responses, our model's macro-F1 score is $0.54$, compared to $0.42$ for ACLR and $0.46$ for BERT. This vital margin demonstrates our model's ability to understand user response patterns and detect misinformation early on. As the number of replies increases, our approach continues to excel. Our model achieves peak performance following 30 responses to the source post. This is crucial for early misinformation detection, as quick identification can prevent the spread of false information. ACLR shows lower performance at early stages, but it gets the best when the entire thread is provided (cf table \ref{tab:results}), showing its lack of ability to perform early misinformation detection. BERT, on the other hand, performs well with few replies as well, but indicating huge limitations in its capacity to utilize the increasing context provided by more replies effectively. Our model's macro-F1 score remains considerably greater than ACLR's and BERT's throughout, especially for fewer replies. This demonstrates both the robustness of our approach and its ability to maintain high accuracy as more user responses are analyzed. In conclusion, \name\ outperforms existing methods in early detection scenarios and provides a reliable tool for combating misinformation.

\subsection{Ablation Study}

The last two rows of Table~\ref{tab:results} present the results of our ablation study, highlighting the contributions of different components of our model. We systematically remove key features of \name\ to determine their impact on overall performance. First, we assess the impact of eliminating the \textit{Q-learning} component. This removal causes a noticeable drop in performance across nearly all metrics. Specifically, the macro-F1 score drops from $0.64$ to $0.58$. The substantial drop emphasizes the importance of our model's Q-learning mechanism, which effectively captures and applies dynamic patterns of user responses to predict misinformation. Next, we investigate the impact of removing the \textit{Text Features} from the model. The performance drop is even more pronounced in this scenario, with the macro-F1 score dropping by more than ten points. This drastic reduction suggests that linguistic cues are critical for understanding and identifying misinformation. The text embeddings will likely capture nuanced contextual information and subtleties in user language required for accurate misinformation detection. Our ablation study results confirm the critical roles of Q-learning and text embeddings in our model. The Q-learning component helps model discourse dynamics and learn from user interactions, both critical for early detection. Text embeddings, conversely, provide a thorough understanding of the linguistic features in posts and responses, which is critical for distinguishing between accurate and false information.

Additionally, we evaluate our model's performance for different claim weights, denoted by $\alpha$. As shown in Table \ref{tab:ablation-alpha}, our experiment results show that setting $\alpha$ to 2 results in the best performance. This weighting strategy recognizes the critical role assertive statements play in determining the integrity of the source post. We observe a substantial decrease in performance when we set it to 1, treating claims and non-claims as equal importance. This finding highlights the importance of emphasizing claims in our model for accurately predicting misinformation. Claims frequently contain more decisive and informative content that aids in determining the veracity of the information, and failing to give them sufficient weight reduces the model's effectiveness. When we increase $\alpha$ to 3, we see a decrease in performance, suggesting that while claims are essential, over-emphasizing them can potentially obscure valuable contextual information provided by non-claims.

\begin{table}[]
\centering
\caption{Performance of \name\ on \dataset\ with varying values of claim weight ($\alpha$).  \label{tab:ablation-alpha}}
\resizebox{0.48\textwidth}{!}{
\begin{tabular}{lccccccc}
\toprule
\multicolumn{1}{c}{\multirow{2}{*}{\textbf{$\alpha$}}} & \multicolumn{3}{c}{\textbf{Non-misinformation}}     & \multicolumn{3}{c}{\textbf{Misinformation}}                 & \multirow{2}{*}{\textbf{M-F1}} \\ \cmidrule{2-7}
\multicolumn{1}{c}{}      & \textbf{P} & \textbf{R} & \textbf{F} & \textbf{P} & \textbf{R} & \textbf{F} &           \\ \midrule
1  & 0.8000  & 0.8000  & 0.8000 & 0.4000  & 0.4000 & 0.4000   & 0.6000    \\
2  & 0.8125 & 0.8667 & 0.8387 & 0.5000  & 0.4000 & 0.4444 & 0.6416 \\
3 & 0.7879	& 0.8667 & 	0.8254 & 0.4286 &	0.3000 & 0.3529 & 0.5892 \\

\bottomrule
\end{tabular}}
\end{table}

\subsection{Error Analysis}

\begin{table*}[!th]
    \centering
    \caption{Error Analysis. Posts with the first five replies (chronological). \label{tab:error_analysis}} 
    \scalebox{0.94}{
    \begin{tabular}{c|p{45em}|c|c|c}
    
    \toprule
    & \textbf{Example} & \textbf{Gold} & \textbf{\name} & \textbf{BERT} \\ 
    \midrule
    $p_1$ &
    \conversation{%
         \parbox{45em}{\textbf{Source Post: Iran turned its most luxurious mall into an emergency hospital, with 3,000 beds to treat coronavirus patients (Meanwhile, the US government keeps escalating its criminal, illegal sanctions on Iran, trying to destroy its economy and prevent it from buying medical equipment).}} \\
            \tikz \draw [arrow1] (0,0.5) -- (0,0) -- (0.4,0); $r_1$: My heart goes out to them. \textit{(\textcolor{purple}{comment}, \textcolor{blue}{non-claim})} \\
            \tikz \draw [arrow1] (0,0.5) -- (0,0) -- (0.4,0); $r_2$: Stay Strong! Keep Safe! Best wish \textit{(\textcolor{purple}{comment}, \textcolor{blue}{non-claim})}\\
            \tikz \draw [arrow1] (0,0.5) -- (0,0) -- (0.4,0); $r_3$: And in USA https://t.co/izvTq3ZLct \textit{(\textcolor{purple}{comment}, \textcolor{blue}{non-claim})}\\
            \tikz \draw [arrow1] (0,0.5) -- (0,0) -- (0.4,0); $r_4$: I have a feeling this is going to turn into real bad karma for the US. \textit{(\textcolor{purple}{comment}, \textcolor{blue}{non-claim})} \\
            \tikz \draw [arrow1] (0,0.5) -- (0,0) -- (0.4,0); $r_5$: Triage in America if the photo is to believed. https://t.co/u3MtLrQ4GZ \textit{(\textcolor{purple}{comment}, \textcolor{blue}{non-claim})} \\
        }
        & FALSE & \textcolor{black}{FALSE} & \textcolor{red}{TRUE} \\ 
  
    \midrule 
    $p_2$ &
    \conversation{%
        \parbox{45em}{\textbf{Source Post: There are still people who aren’t taking this \#coronavirus outbreak seriously. Remember, it’s not just about you. Spare a thought for family \& loved ones, especially the elderly. You may be young \& healthy but there are those whose immune system is compromised. Don’t be selfish!}} \\
        
            \tikz \draw [arrow1] (0,0.5) -- (0,0) -- (0.4,0); $r_1$: Subhanallah may Allah Almighty save us all from this calamity.........Amin \textit{(\textcolor{purple}{comment}, \textcolor{blue}{non-claim})} \\
            \tikz \draw [arrow1] (0,0.5) -- (0,0) -- (0.4,0); $r_2$: Indeed shiekh, i agree. \textit{(\textcolor{purple}{support}, \textcolor{blue}{non-claim})}\\
            \tikz \draw [arrow1] (0,0.5) -- (0,0) -- (0.4,0); \parbox{43em}{$r_3$: Is rather unfortunate and sad and this people happen to be people that people looks upto, rely on and respect their words. \textit{(\textcolor{purple}{support}, \textcolor{blue}{non-claim})}} \\
            \tikz \draw [arrow1] (0,0.5) -- (0,0) -- (0.4,0); \parbox{43em}{$r_4$: tell them pls sir. Our scholar here are telling us \#coronavirus its lie and a western propaganda we shouldn't accept it. Shall we sir? \textit{(\textcolor{purple}{query}, \textcolor{blue}{non-claim})}} \\
                \hspace{5pt} \tikz \draw [arrow1] (0,0.5) -- (0,0) -- (0.4,0); \parbox{41em}{$r_5$: God bless you! I watched a video saying this yesterday. I was just like wow! They think this virus is a joke. \textit{(\textcolor{purple}{support},\textcolor{blue}{claim})}} \\
        } 
        & FALSE	& \textcolor{red}{TRUE} & \textcolor{black}{FALSE} \\     
    
    \midrule
    $p_3$ &
    \conversation{%
        \parbox{45em}{\textbf{Source Post: As of yesterday Canada has completed 1.5 times more coronavirus tests than US. On a per capita basis? 13 times }} 
        \\
            \tikz \draw [arrow1] (0,0.5) -- (0,0) -- (0.4,0); \parbox{43em}{$r_1$: We should not really be comparing our health care system or our Federal Government to the US. It’s not a level playing field. \textit{(\textcolor{purple}{deny}, \textcolor{blue}{non-claim})}} \\

                \hspace{5pt} \tikz \draw [arrow1] (0,0.5) -- (0,0) -- (0.4,0);  \parbox{41em}{$r_2$: Not sure of the intent of your tweet. The table is tilted in favour of? \textit{(\textcolor{purple}{query}, \textcolor{blue}{non-claim})}} \\

                    \hspace{10pt} \tikz \draw [arrow1] (0,0.5) -- (0,0) -- (0.4,0);  \parbox{39em}{$r_3$: In favour of Canadians. We do not have to worry about the cost of going to the hospital. \textit{(\textcolor{purple}{comment}, \textcolor{blue}{non-claim})}} \\

                        \hspace{15pt} \tikz \draw [arrow1] (0,0.5) -- (0,0) -- (0.4,0);  \parbox{40em}{$r_4$:  The other half of my position is that our Federal government is...functioning... Dr. Theresa Tam has a clear and consistent message.  \textit{(\textcolor{purple}{query}, \textcolor{blue}{non-claim})}} \\

                            \hspace{20pt} \tikz \draw [arrow1] (0,0.5) -- (0,0) -- (0.4,0);  \parbox{40em}{$r_5$: Source of the info? \textit{(\textcolor{purple}{query}, \textcolor{blue}{non-claim})} } \\
        }
        & TRUE & \textcolor{black}{TRUE} & \textcolor{red}{FALSE} \\ 
         
    \bottomrule

    \end{tabular}
    }
    \label{tab:descriptions}
\end{table*}

To reaffirm our point, detecting misinformation on online social media is finicky due to the profoundly subjective nature of claims mentioned in them and the folksiness of these platforms. Glancing at Table \ref{tab:results}, we can see that all the systems, including ours, inevitably make mistakes in identifying these posts as misinformation as the F1-score for the true label is much lesser than that of the false. To further comprehend the performance of our proposed model \name, now we strive to qualitatively analyze the errors committed by our proposed model, \name. Table \ref{tab:error_analysis} highlights randomly sampled error cases from our test dataset, along with their gold labels and predictions from \name. In addition, for comparison, we furnish the predictions from the best-performing baseline, BERT. 

Consider that first post ($p_1$) about Iran converting a mall into an emergency hospital and the impact of U.S. sanctions; our model successfully identified the content as `False' (not misinformation), aligning with the gold label. This accuracy suggests that our model robustly understands global contexts and can effectively differentiate between emotionally charged language and factual information. It leveraged thread information to contextualize the statement within a broader discussion, enhancing its judgment accuracy. In contrast, the baseline model, BERT, incorrectly labeled the post as `True' (misinformation), indicating potential weaknesses. This error could stem from the baseline's insufficient handling of geopolitical nuances and crisis-related content. Integrating conversation thread into the analysis helped our model to provide more contextually aware assessments. 

In the second example, the post underscores the seriousness of the coronavirus outbreak and the need for collective responsibility; our model incorrectly classified the post as `True' (misinformation), which diverges from the gold label `False' (not misinformation). This error suggests our model may have misinterpreted the post's underlying intent or factual content, possibly due to the emotionally charged call to action or misunderstanding contextual nuances in the user comments. On the other hand, the baseline model BERT accurately classified the post as `False' (not misinformation), showing better alignment with the gold label. This indicates that the BERT model may have a more effective mechanism for interpreting the factual nature of health-related advisories. Despite being an error relative to the gold label, there is a nuanced aspect to consider that suggests some potential benefits of such false positives, with varied user comments. Public health messaging requires high accuracy and sensitivity, especially during a crisis. A model that errs on the side of caution by marking potentially exaggerated or emotionally charged statements as misinformation might help maintain a stringent check on the quality and reliability of information spread during crucial times. This cautious approach ensures that only well-substantiated, clear, and responsibly communicated messages gain widespread traction, thus preventing panic or misinformation. 

The last example, about Canada's coronavirus testing compared to the US, we observe that our model correctly classified the claim as `True,' aligning with the gold label, while the baseline model incorrectly flagged it as `False.' Comments like \textit{`We should not really be comparing our health care system or our Federal Government to the US'} and \textit{`Not sure of the intent of your tweet}' might suggest doubt or denial in the conversation thread that challenges the factual accuracy of the initial claim made in the source post. Even though labeled as non-claims, these comments have aided our model in questioning the post's veracity. It indicates that our model may better handle the context or use more reliable external sources to affirm the truthfulness of statistical claims, maintaining its accuracy despite mixed sentiments in the comments.

\section{Discussion}
We analyze our proposed framework and baseline systems based on the following comprehensive set of questions to ensure a thorough evaluation and identify areas for improvement.

\textbf{\textit{How accurately do models with just the source post predict?}} 
Misinformation detection solely based on the source post is challenging due to the limited context it provides. Often short and lacking detailed information, social media posts may not convey the complete picture. Despite this limitation, BERT emerges as a standout performer among all other systems in the first group. Despite leveraging additional contextual cues from the conversation thread, BERT demonstrates remarkable predictive capabilities. BERT`s success can be attributed to its ability to capture nuanced semantic information from the source post. By encoding the text into dense embeddings and leveraging its pre-trained knowledge, BERT effectively discerns misinformation patterns within the source content. It is worth noting that while BERT excels in this setup, other baselines such as LSTM and RoBERTa also exhibit commendable performance based solely on the source post. This suggests that while additional context from conversation threads can be beneficial, sophisticated language models like BERT, LSTM, and RoBERTa can extract meaningful information even from standalone posts.

\textbf{\textit{Does the addition of replies enhance BERT's performance in misinformation detection?}}
In the initial evaluation using solely the source post, BERT demonstrates remarkable performance, indicating its proficiency in identifying misinformation without additional contextual cues. However, we integrate conversation replies into the BERT model in three distinct configurations to ensure equitable comparisons. Firstly, we augment all the replies categorized as `support,' `deny,' or `query' alongside the source post. Secondly, we include replies labeled as claims. Finally, we incorporate replies classified as claims or one of the stances mentioned earlier. Among these configurations, the second setup yields the most favorable results, where only replies categorized as claims are included along with the source post. This observation underscores the significance of claims within conversation threads as valuable indicators of misinformation. Interestingly, the mere addition of these replies only significantly enhances the performance of BERT compared to its performance solely based on the source post. This suggests that while claims are indeed informative components within conversation threads, they do not substantially augment BERT`s ability to detect misinformation when integrated alongside the source post. Other setups signify the inability of the BERT to deal with extraneous texts.  

\textbf{\textit{Does the conversation thread structure help in detection?}}
Integrating conversation threads alongside the source post provides supplementary context for misinformation detection. However, the actual impact of conversation threads on the detection process varies. While conversation threads offer supplementary context, sophisticated language models like BERT remain potent tools for misinformation detection, capable of extracting meaningful insights even from standalone posts. Balancing the integration of conversation threads with advanced language models presents an ongoing challenge in pursuing more accurate misinformation detection methodologies. On the other hand,  the addition of conversation thread, stance labels, and claim information appropriately guides \name\ in identifying the false information.

\textbf{\textit{Are LLMs capable of detecting misinformation?}}
Examining the results of GPT-4 experiments offers valuable insights into the dynamics of misinformation detection using LLMs. The GPT-4 experiment emphasized the importance of incorporating conversation threads, revealing their potential to enhance detection accuracy. While GPT-4 performed highly without explicitly considering conversation threads, the GPT-4 experiment demonstrated notable improvements when replies were included. These findings highlight the significance of advanced models and contextual cues in misinformation detection. Achieving optimal results in this field requires a delicate balance between leveraging sophisticated models and integrating supplementary contextual information from conversation threads.

\section{Conclusion}
In conclusion, our research makes a substantial contribution to the growing field of combating the widespread spread of misinformation on social media platforms. We presented a novel framework for early misinformation detection that combines crowd intelligence with sophisticated reinforcement learning mechanisms. By incorporating user stances and claims, our model can quickly detect and flag potentially deceptive content, fostering a sense of community and facilitating informed public debate. We demonstrated how using user stances and assertions on social media posts can help to make better decisions about the truth or falsity of posts at an early stage. Furthermore, improved efficiency in misinformation detection may help manual fact-checkers prioritize fact-checking more quickly. One of the major obstacles to misinformation detection in online social media is a lack of a sufficiently comprehensive annotated dataset. As a result, we created \dataset, a Twitter corpus of manually annotated conversation threads for misinformation detection, complete with replies, stances, and claim labels. We developed a unified architecture, \name\, which outperformed existing state-of-the-art systems and demonstrated how the abetting RL could be used to understand crowd intelligence better. The results showed that our proposed model outperformed the best-performing baselines by $\geq$ $0.4$\% in the macro-F1 score. We acknowledge that we have focused on English in this work; therefore, in future work, we will seek to develop misinformation detection models for other languages, particularly low-resource languages. In the future, we plan to expand the dataset and improve the performance of the minority class. Finally, we intend to expand our efforts to include multimodality, like images, memes, URLs, etc.

\bibliographystyle{IEEEtran}
\bibliography{bib}


\section{Biography Section}
\vspace{-33pt}
\begin{IEEEbiographynophoto}{Megha Sundriyal} is a PhD student at IIIT Delhi, India. Her primary research interests are Natural Language Processing and Online Social Media, where she aims to use computational techniques to examine and comprehend human language and social interactions in online spaces.
\vspace{-33pt}
\end{IEEEbiographynophoto}
\begin{IEEEbiographynophoto}{Harshit Choudhary} is a BTech student at IIIT Delhi, India. With a keen interest in Natural Language Programming, he is dedicated to exploring the intricacies of human language through the lens of technology.
\end{IEEEbiographynophoto}
\vspace{-33pt}
\begin{IEEEbiographynophoto}{Tanmoy Chakraborty} is an Associate Professor at IIT Delhi, India. Previously, he was a faculty member at IIIT Delhi, India. His primary research interests include Social Computing, Graph Mining, and Natural Language Processing. He is a senior IEEE member.
\end{IEEEbiographynophoto}
\vspace{-33pt}
\begin{IEEEbiographynophoto}{Md Shad Akhtar} is an Assistant Professor at IIIT Delhi, India. His main area of research is Natural Language Processing, focusing on affective analysis and computational social systems. He completed his PhD from IIT Patna, India.
\end{IEEEbiographynophoto}
 



\section*{Supplementary}
\subsection{Annotation Guidelines}
\label{sec:annotations}
\subsubsection{Misinformation Annotation} We define misinformation as any information that circulates on social media but is later proven false or inaccurate. Source posts are labeled with 1 for misinformation and 0 for non-misinformation.

\begin{itemize}
    \item \textbf{Non-misinformation:} A post is considered non-misinformation if it can be verified by credible sources such as Wikipedia, published reports, or fact-checking websites like Snopes or Politifact.
    
    \textit{Example: `Suspects in \#CharlieHebdo attack spoke to police by phone and said they wanted to die as martyrs, says French MP.'}

    \item \textbf{Misinformation:} A post is considered misinformation if it contains information verified as false by credible sources.

    \textit{Example: `If you told me on New Year that within 3 months of 2020 -- Kobe Bryant is dead. We’d be quarantined from COVID-19. Schools and sports are shut down. Tom Brady is a Tampa Bay Buccaneer. I would’ve legitimately thought the world was ending.'}
\end{itemize} 

\subsubsection{Stance Annotation} Stance is the specific position or attitude that a statement takes with respect to some target. In this work, we define a reply's stance as its position in relation to the source post. Thus, every reply can be labeled with one of four stances -- support, deny, query, or comment.

\begin{itemize}
    \item \textbf{Support}: A reply that holds up the source tweet by promoting the validity or importance of the source tweet with phrases like `Thank you,' `Well done,' etc, or providing facts and statistics that give more weight to the source tweet.

    \textit{Example: `Yes, The rising COVID cases in the country indeed demand immediate action by the government'} in response to the source post, \textit{`Rising COVID cases demand immediate action.'}

    \item \textbf{Deny}: A reply that refutes the source tweet by directly negating the source post or providing facts and statistics that directly contradict it.

    \textit{Example: `It is because Japan has the world's largest old age population.']} is annotated as deny for the source tweet \textit{`Due to lack of management and Healthcare Facilities, a lot of people are dying in Japan.'}

    \item \textbf{Query}: A reply questioning something about the source tweet, thereby bringing a neutral opinion towards the source tweet due to lack of knowledge.

    \textit{Example: `How is the lab running now? Full capacity or at reduced function.'}
    
    \item \textbf{Comment}: A comment stance indicates that the user's reply does not take a specific position on the original post. 

    \textit{Example: `@KiranKS I would like to know, too.'}
    
\end{itemize}

\subsubsection{Claim Annotation} For claim annotations of the replies, we utilize the comprehensive annotation guidelines provided by Gupta et al. \cite{gupta2021lesa}. A claim is a statement that asserts a belief or fact that can be challenged or supported by evidence, whereas a non-claim is a statement that makes no assertions that need to be clarified. Examples of claims and non-claims are as follows:

\textbf{Claim Examples}
\begin{itemize}
    \item \textit{`Covid's first case was reported on 21 December 2019 and till date approximately 50,000 people have died.'}
    \item \textit{`Italy does not have the world's second-largest old age population.'}
\end{itemize}

\textbf{Non-claim Examples}
\begin{itemize}
    \item \textit{`If you think drinking disinfectants will cure \#Covid 19 , you deserve death \#trump'}
    \item \textit{`Do disinfectants really cure Corona?'}
\end{itemize}

\begin{table*}[!ht]
\centering
\caption{Dataset statistics showing stance labels within the conversational threads. \label{tab:dataset-stance-stats}}
\begin{tabular}{c|cccc|cccc|cccc}
\toprule
\textbf{} & \multicolumn{4}{c|}{\textbf{\dataset}} & \multicolumn{4}{c|}{\textbf{SemEval$_{Twitter}$}} & \multicolumn{4}{c}{\textbf{SemEval$_{Reddit}$}} \\ \cmidrule{2-13} 

\textbf{} & \textbf{Support}  & \textbf{Deny}  & \textbf{Query} & \textbf{Comment} & \textbf{Support}  & \textbf{Deny}  & \textbf{Query} & \textbf{Comment} & \textbf{Support}  & \textbf{Deny}  & \textbf{Query} & \textbf{Comment}  \\ \midrule
Train     & 1382   & 432 & 911   & 7819   &   1004 & 415 & 464 & 3685 & 23 & 45 & 51 & 996  \\
Test      & 409   & 182    & 302      & 2959   &   141 &  92 & 62 &  771 & 16 & 10 & 31 & 633 
 \\ \bottomrule
\end{tabular}
\end{table*}

\subsection{Additional Dataset Statistics}
Table \ref{tab:dataset-stance-stats} presents a comprehensive summary of dataset statistics for stance labels in conversational threads across three datasets: \dataset\ (our dataset), SemEval$_{twitter}$, and SemEval$_{reddit}$. The stance labels are classified as Support, Deny, Query, and Comment. The \dataset\ dataset consists of 1382 instances of Support in the training subset, 432 instances of Deny, 911 instances of Query, and 7819 instances of Comment. In the testing subset, there are 409 instances of Support, 182 instances of Deny, 302 instances of Query, and 2959 instances of Comment. Overall, \dataset\ dataset has the highest number of instances, particularly in the Comment category. It is then followed by \textit{SemEval$_{twitter}$} dataset and \textit{SemEval$_{reddit}$}. This statistical distribution offers insights into the balance and composition of the datasets across various stance labels.

\subsection{Additional Experimental Results}
As mentioned in the main paper, we also perform our experiments on SemEval-2029 task 7 datasets \cite{gorrell2019semeval}. Table \ref{tab:results-semeval-twitter} presents the performance of our model and baseline systems SemEval$_{Twitter}$ dataset. Our proposed system \name\ demonstrates exceptional performance in accurately identifying misinformation and obtaining a high recall and precision. It attains a Macro-F1 score of 0.5317. For the SemEval$_{Reddit}$ dataset, the results are shown in Table \ref{tab:results-semeval-reddit}. \name\ performs well, especially for the misinformation class, achieving the highest macro-F1 score of 0.5192. EventAI \cite{eventai2019semeval} achieves high recall for the Non-misinformation class but zero performance for the misinformation class, resulting in a low macro-F1 score of 0.2857. GCN \cite{bacciu2020gentle} exhibits a similar performance to EventAI with a macro-F1 score of 0.3056. ACLR  \cite{lin2022detect} faces the same issues as other models with low overall performance, yielding a macro-F1 score of 0.3056. RPL \cite{lin2023zero} and GMVCN \cite{wu2024graph} show balanced performance with macro-F1 scores of 0.4318.

\begin{table*}[ht]
\centering
\caption{Experimental results of \name\ on the  SemEval$_{Twitter}$ dataset. \label{tab:results-semeval-twitter}}
\begin{tabular}{lccccccc}
\toprule
\multicolumn{1}{c}{\multirow{2}{*}{\textbf{Model}}} & \multicolumn{3}{c}{\textbf{Non-misinformation}}                & \multicolumn{3}{c}{\textbf{Misinformation}}                 & \multirow{2}{*}{\textbf{Macro-F1}} \\ \cmidrule{2-7}
\multicolumn{1}{c}{}                                & \textbf{Pecision} & \textbf{Recall} & \textbf{F1} & \textbf{Pecision} & \textbf{Recall} & \textbf{F1} &                                    \\ \midrule
LSTM           & 0.5750                & 0.7188               & 0.6389               & 0.4375               & 0.2917               & 0.3500                 & 0.4944               \\
RNN            & 1.0000                    & 0.0625               & 0.1176               & 0.4444               & 1.0000                    & 0.6154               & 0.3665               \\
BERT           & 0.5652               & 0.4062               & 0.4727               & 0.4242               & 0.5833               & 0.4912               & 0.4820                \\
RoBERTa        & 0.5556               & 0.3125               & 0.4000                 & 0.4211               & 0.6667               & 0.5161               & 0.4581               \\
\midrule
EventAI & 0.7857  & 0.6875  &  0.7333 & 0.6429  &  0.7500   &  0.6923 & 0.7128 \\ 
GCN     & 0.5100                 & 0.6200                 & 0.5600                 & 0.2900                 & 0.2100                 & 0.2400                 & 0.4000                  \\ 

ACLR   & 0.6667               & 0.0625               & 0.1143               & 0.434                & 0.9583               & 0.5974               & 0.3558               \\
RPL      & 0.5938 &	0.5938 &	0.5938	& 0.4583 &	0.4583	& 0.4583 &	0.5260 \\
GMVCN          & 0.4400                 & 0.1200                 & 0.2000                  & 0.400                  & 0.7900                 & 0.5400                 & 0.3700                 \\
\midrule

\name     & 0.6122               & 0.9375               & 0.7407               & 0.7143               & 0.2083               & 0.3226               & 0.5317  \\            
 \bottomrule
\end{tabular}
\end{table*}

\begin{table*}[]
\centering
\caption{Experimental results of \name\ on the  SemEval$_{Reddit}$ dataset. \label{tab:results-semeval-reddit}}
\begin{tabular}{lccccccc}
\toprule
\textbf{}      & \multicolumn{3}{c}{\textbf{Non-misinformation}}                 & \multicolumn{3}{c}{\textbf{Misinformation}}                  & \multicolumn{1}{l}{\multirow{2}{*}{\textbf{Macro-F1}}} \\ \cmidrule{2-7}

\textbf{Model} & \textbf{Precision} & \textbf{Recall} & \textbf{F1} & \textbf{Precision} & \textbf{Recall} & \textbf{F1} & \multicolumn{1}{l}{}                                   \\
\midrule
LSTM           & 0.4400               & 1.0000               & 0.6111      & 0.0000                 & 0.0000               & 0.0000           & 0.3056                                                 \\
RNN            & 0.4400               & 1.0000               & 0.6111      & 0.0000                  & 0.0000               & 0.0000           & 0.3056                                                 \\
BERT           & 0.4545             & 0.9091          & 0.6061      & 0.6667             & 0.1429          & 0.2353      & 0.4207                                                 \\
RoBERTa        & 0.4286             & 0.8182          & 0.5625      & 0.5000                & 0.1429          & 0.2222      & 0.3924                                                 \\
\midrule
EventAI  & 0.4167    & 0.9091  &   0.5714 & 0.0000  &   0.0000  &   0.0000 &  0.2857 \\
GCN    & 0.44               & 1.0000               & 0.6111      & 0.0000                  & 0.0000               & 0.0000           & 0.3056                                                 \\
ACLR      & 0.4400               & 1.0000               & 0.6111      & 0.0000                  & 0.0000              & 0.0000           & 0.3056                                                 \\
RPL     & 0.3636             & 0.3636          & 0.3636      & 0.5000                & 0.5000             & 0.5000         & 0.4318                                                 \\
GMVCN          & 0.3636             & 0.3636          & 0.3636      & 0.5000                & 0.5000             & 0.5000         & 0.4318                                                 \\
\midrule
\name     & 0.4667             & 0.6364          & 0.5385      & 0.6000                & 0.4286          & 0.5000         & 0.5192   
\\
\bottomrule
\end{tabular}
\end{table*}

\end{document}